\documentclass{article}

\usepackage[preprint]{neurips_2025}

\usepackage[utf8]{inputenc}
\usepackage[T1]{fontenc}
\usepackage{hyperref}
\usepackage{url}
\usepackage{booktabs}
\usepackage{amsfonts}
\usepackage{nicefrac}
\usepackage{microtype}
\usepackage{xcolor}
\usepackage{soul}
\usepackage{amsmath,amssymb,graphicx,color,algorithm,subfig, algpseudocode,booktabs,bm,relsize,enumitem,multirow,amsthm,epsfig,caption}
\usepackage[export]{adjustbox}
\usepackage{lscape}
\usepackage{multirow}
\usepackage{graphicx}
\usepackage{booktabs}
\usepackage{makecell}

\usepackage{microtype}
\usepackage{graphicx}
\usepackage{hyperref}

\usepackage{amsmath}
\usepackage{amssymb}
\usepackage{mathtools}
\usepackage{amsthm}
\usepackage{wrapfig}

\usepackage[capitalise,noabbrev,nameinlink]{cleveref}
\creflabelformat{equation}{#2\textup{#1}#3}
\creflabelformat{section}{#2\textup{#1}#3}
\creflabelformat{appendix}{#2\textup{#1}#3}
\creflabelformat{figure}{#2\textup{#1}#3}
\crefname{algocf}{Algorithm}{Algorithms}
\Crefname{algocf}{Algorithm}{Algorithms}

\definecolor{DarkBlue}{rgb}{0,0.08,0.45}

\hypersetup{
colorlinks=true,
linkcolor=DarkBlue,
citecolor=DarkBlue,
filecolor=DarkBlue,
urlcolor=DarkBlue}

\usepackage{amsmath,amssymb,graphicx,color,algorithm,subfig, algpseudocode,booktabs,bm,relsize,enumitem,multirow,amsthm,epsfig,caption}
\usepackage[export]{adjustbox}
\usepackage{lscape}
\usepackage{multirow}
\usepackage{graphicx}
\usepackage{booktabs}
\usepackage{makecell}

\newsavebox{\measurebox}

\definecolor{cornellred}{rgb}{0.7, 0.11, 0.11}
\definecolor{cadmiumgreen}{rgb}{0.0, 0.42, 0.24}
\definecolor{aliceblue}{rgb}{0.91, 0.94, 0.97}
\definecolor{darkblue}{rgb}{0.83, 0.89, 0.97}
\definecolor{Red7}{rgb}{0.941, 0.243, 0.243}
\definecolor{Green7}{RGB}{55, 178, 77}
\definecolor{Blue9}{rgb}{0.098,0.3,0.9}

\usepackage{pifont}

\sethlcolor{aliceblue}

\title{FastTD3: Simple, Fast, and Capable 
\\ Reinforcement Learning for Humanoid Control}

\author{
  Younggyo Seo$^1$ \; Carmelo Sferrazza$^1$ \; Haoran Geng$^1$ \\  \textbf{Michal Nauman}$^{1,2}$ \; \textbf{Zhao-Heng Yin}$^1$ \; \textbf{Pieter Abbeel}$^1$ \\
  $^1$University of California, Berkeley \; $^2$University of Warsaw \\[0.25cm]
\normalsize{\textbf{\href{https://younggyo.me/fast_td3}{\texttt{https://younggyo.me/fast\_td3}}}} \\
}

\begin{document}

\maketitle

\vspace{-0.2in}

\begin{abstract}
  Reinforcement learning (RL) has driven significant progress in robotics, but its complexity and long training times remain major bottlenecks.
  In this report, we introduce FastTD3, a simple, fast, and capable RL algorithm that significantly speeds up training for humanoid robots in popular suites such as HumanoidBench, IsaacLab, and MuJoCo Playground.
  Our recipe is remarkably simple: we train an off-policy TD3 agent with several modifications -- parallel simulation, large-batch updates, a distributional critic, and carefully tuned hyperparameters.
  FastTD3 solves a range of HumanoidBench tasks in under 3 hours on a single A100 GPU, while remaining stable during training.
  We also provide a lightweight and easy-to-use implementation of FastTD3 to accelerate RL research in robotics.
\end{abstract}

\begin{figure*}[h]
  \vspace{-0.05in}
  \center
  \makebox[\textwidth]{ \includegraphics[width=0.85\paperwidth]{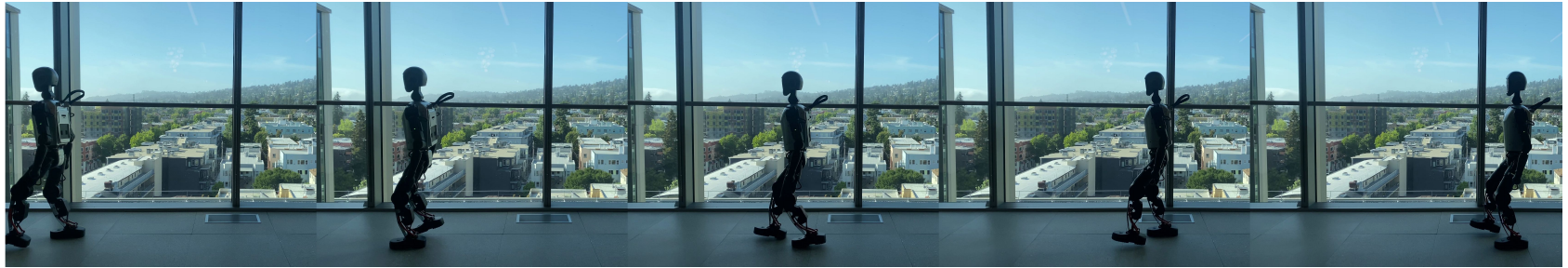}}
  \caption{\textbf{Sim-to-real reinforcement learning with FastTD3.} We successfully transfer the FastTD3 policy trained in MuJoCo Playground \citep{zakka2025mujoco} to Booster T1.
  This represents, to the best of our knowledge, the first documented instance of a successful deployment of an off-policy RL policy on a full-size humanoid robot in the real world.
  See our \href{https://younggyo.me/fast_td3}{project page} for videos.
  }
  \label{fig:sim2real}
\end{figure*}

\vspace{-0.1in}

\section{Introduction}
Reinforcement learning (RL) has been a key driver behind recent successes in robotics, enabling the successful transfer of robust simulation policies to real-world environments \citep{hwangbo2019learning,kaufmann2023champion}. However, progress is often bottlenecked by slow training times in complex tasks. For example, in the recently proposed benchmark HumanoidBench \citep{sferrazza2024humanoidbench}, even state-of-the-art RL algorithms failed to solve many tasks after 48 hours of training.

This slow training remains a major bottleneck for practitioners aiming to unlock new behaviors in humanoid robots using RL. In particular, the iterative nature of reward design in robotics -- where multiple rounds of reward shaping and policy retraining are often necessary -- demands RL algorithms that are not only \textit{capable} but also significantly \textit{faster}. These algorithms must support rapid iteration and reliably solve tasks when given well-designed rewards.

\begin{figure*} [t!] \centering
\includegraphics[width=1.0\textwidth]{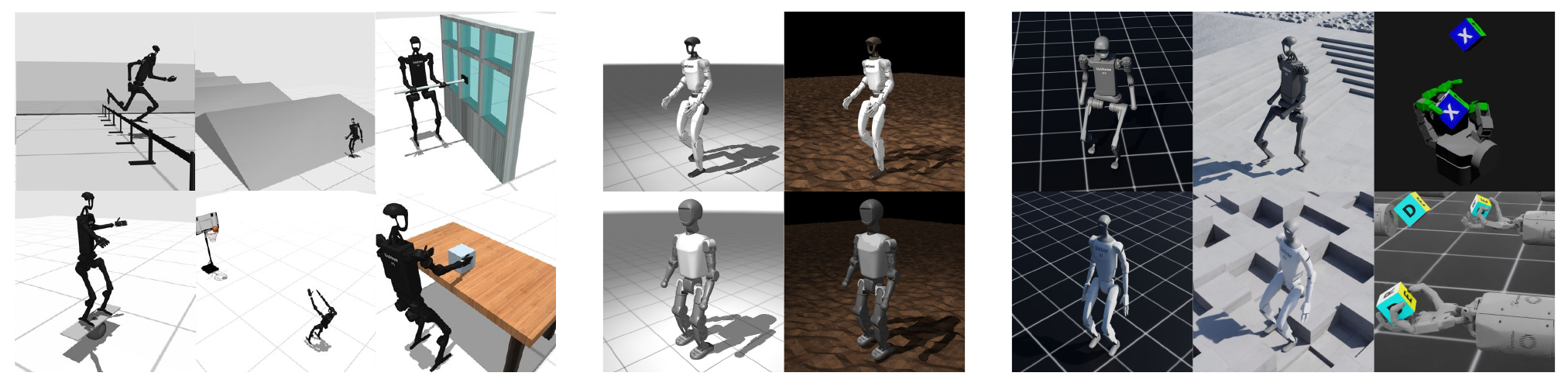}
\caption{\textbf{Tasks.} We consider a range of challenging humanoid and dexterous control tasks from HumanoidBench (left), MuJoCo Playground (middle), and  IsaacLab (right).}
\label{fig:tasks}
\end{figure*}

\begin{figure*}[h]
\centering
\vspace{-0.15in}
\includegraphics[width=1.0\linewidth]{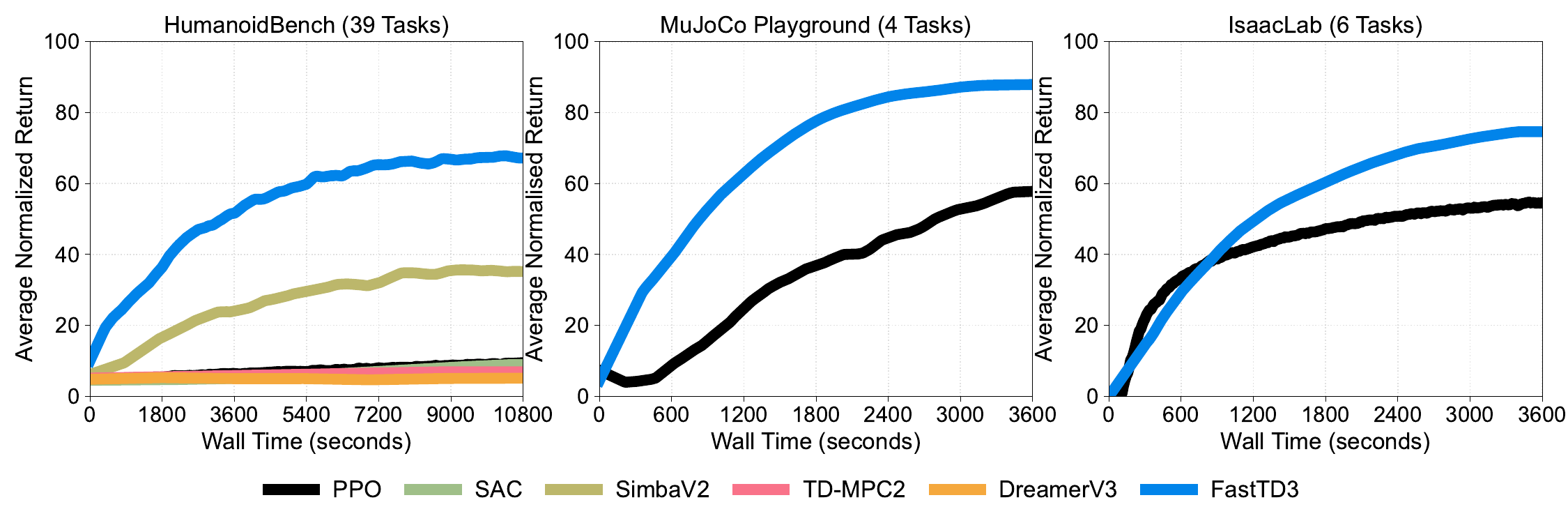}
\vspace{-0.075in}
\caption{\textbf{Summary of results.} FastTD3 is a simple, fast, and capable RL algorithm that significantly speeds up training for humanoid robots on tasks from popular suites such as HumanoidBench \citep{sferrazza2024humanoidbench}, IsaacLab \citep{mittal2023orbit}, and MuJoCo Playground \citep{zakka2025mujoco}.
To accelerate RL research in robotics, we provide an easy-to-use open-source implementation of FastTD3, enabling users to easily reproduce these results or build upon our work.
}
\label{fig:aggregate}
\end{figure*}

For that purpose, practitioners have mostly used Proximal Policy Optimization (PPO; \citealt{schulman2017proximal}) for training deployable policies in simulation, as PPO learns behaviors very fast with massively parallel simulation \citep{heess2017emergence,hwangbo2019learning}.
However, PPO is an on-policy algorithm and not sample-efficient, making it difficult to fine-tune it during real-world deployment or initialize training with demonstrations \citep{hester2018deep}.

Meanwhile, recent off-policy RL research has made significant progress in improving sample-efficiency \citep{d'oro2023sampleefficient,nauman2024bigger,hansen2023td}.
However, this line of work often suffers from increased algorithmic complexity and long wall-clock training times, making them difficult to be widely used for learning deployable policies in robotics.

On the other hand, Parallel Q-Learning (PQL; \citealt{li2023parallel}) have shown that off-policy RL can be both fast and sample-efficient by scaling up through massively parallel simulation, large batch sizes, and a distributional critic \citep{bellemare2017distributional}. 
However, the core contribution of PQL -- its use of asynchronous parallel processes that cuts wall-clock time -- unfortunately comes with high implementation complexity, which has hindered its widespread adoption.
In this work, we do not aim to claim novelty over PQL, but rather focus on developing a simple yet highly-optimized algorithm without asynchronous processes, highlighting its effectiveness on popular humanoid control suites, and providing the easy-to-use implementation to accelerate future RL research on robotics.

\vspace{-0.015in}
\paragraph{FastTD3} We introduce FastTD3, a simple, fast, and capable RL algorithm that significantly speeds up training for humanoid robots on tasks from popular suites such as HumanoidBench \citep{sferrazza2024humanoidbench}, IsaacLab \citep{mittal2023orbit}, and MuJoCo Playground \citep{zakka2025mujoco}.
Our recipe is remarkably simple: by training an off-policy TD3 agent \citep{fujimoto2018addressing} with parallel simulation, large-batch updates, a distributional critic \citep{bellemare2017distributional}, and carefully tuned hyperparameters, FastTD3 solves a range of HumanoidBench tasks in under 3 hours on a single GPU.
Compared to PPO \citep{schulman2017proximal}, FastTD3 trains humanoid locomotion policies faster in IsaacLab and MuJoCo Playground, particularly on rough terrain with domain randomization.

\clearpage

\paragraph{Open-source implementation} We provide an open-source implementation of FastTD3 based on PyTorch \citep{paszke2019pytorch} -- a lightweight codebase that enables users to easily build new ideas on top of FastTD3.
Our implementation is easy-to-install and versatile -- users can easily train FastTD3 agents on HumanoidBench, IsaacLab, and MuJoCo Playground, and the codebase also supports several user-friendly features, such as preconfigured hyperparameters, rendering support, logging, and loading checkpoints.
We note that our work is orthogonal to latest RL research, such as SR-SAC \citep{d'oro2023sampleefficient}, BBF \citep{schwarzer2023bigger}, BRO \citep{nauman2024bigger}, Simba \citep{lee2024simba}, NaP \citep{lyle2024normalization}, TDMPC2 \citep{hansen2023td}, TDMPBC \citep{zhuang2025tdmpbc}, SimbaV2 \citep{lee2025hyperspherical}, MAD-TD \citep{voelcker2024mad}, and MR.Q \citep{fujimoto2025towards}, we expect various improvements in these works to be also useful when incorporated into FastTD3.

Our key contributions can be summarized as follows:
\begin{itemize}
    \item We introduce FastTD3, a simple, fast and capable RL algorithm that solves a variety of locomotion and manipulation tasks that prior RL algorithms take tens of hours to complete or fail to solve.
    We demonstrate that this performance can be achieved using a remarkably simple recipe: training a TD3 agent \citep{fujimoto2018addressing} with large-batch updates, parallel simulation, distributional RL, and well-tuned hyperparameters.
    \item We provide experimental results that show the effectiveness of the various design choices.
    \item We release an easy-to-use open-source implementation of FastTD3 to accelerate RL research on robotics. This implementation supports popular suites such as HumanoidBench \citep{sferrazza2024humanoidbench}, IsaacLab \citep{mittal2023orbit}, and MuJoCo Playground \citep{zakka2025mujoco}.
\end{itemize}

\begin{figure*} [t!] \centering
\vspace{-0.05in}
\includegraphics[width=0.99\textwidth]{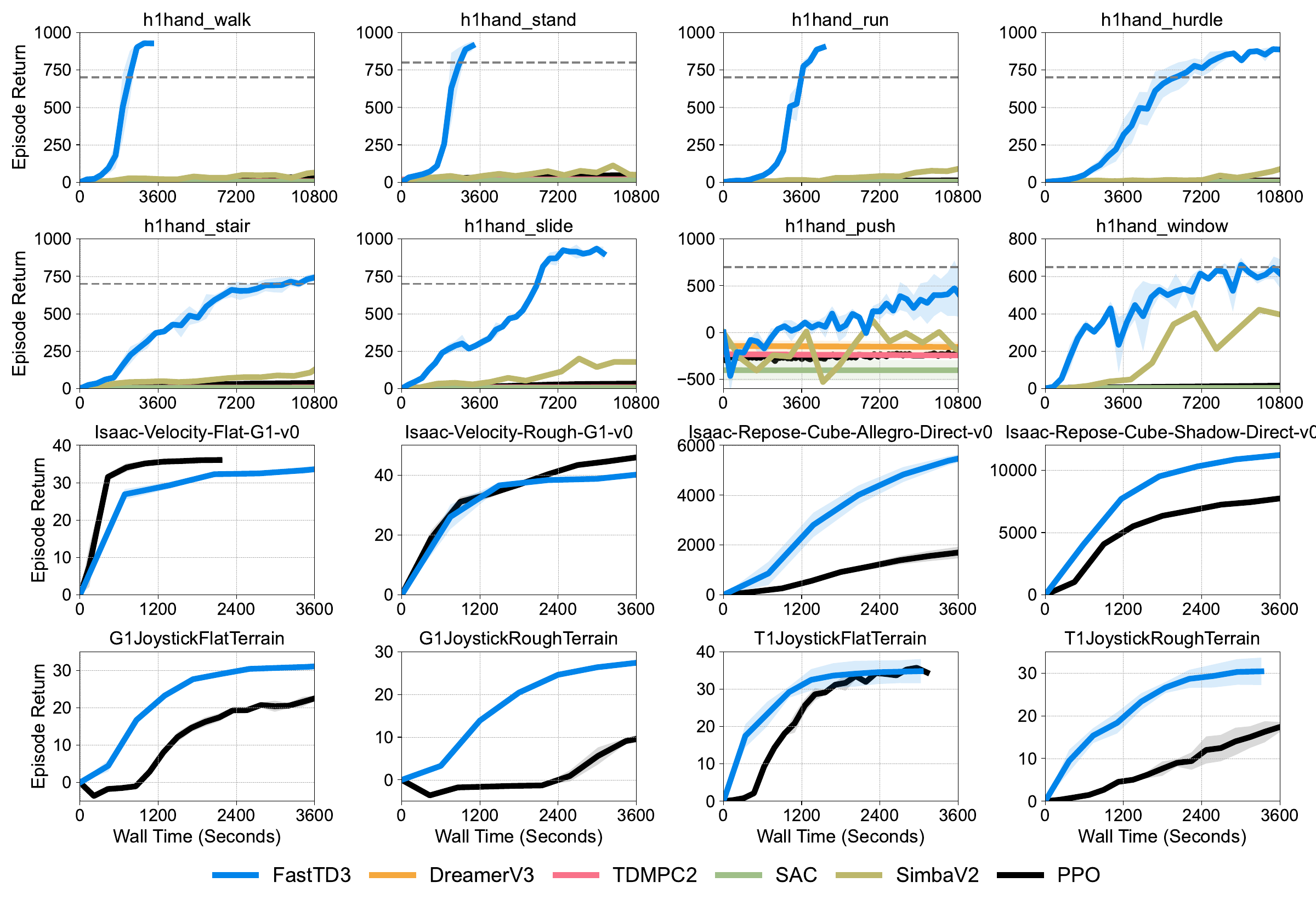}
\vspace{-0.05in}
\caption{\textbf{Results on a selected set of tasks.} Learning curves on selected individual tasks from HumanoidBench (first two rows), IsaacLab (third row), and MuJoCo Playground (fourth row). The solid line and shaded regions represent the mean and standard deviation across three runs. The dashed lines indicate success thresholds in HumanoidBench tasks.}
\label{fig:aggregate_total}
\end{figure*}

\vspace{-0.05in}
\section{FastTD3: Simple, Fast, Capable RL for Humanoid Control}
\vspace{-0.05in}

FastTD3 is a high-performance variant of the Twin Delayed Deep Deterministic Policy Gradient (TD3; \citealt{fujimoto2018addressing}) algorithm, optimized for complex robotics tasks. These optimizations are based on the observations of \citet{li2023parallel} that found parallel simulation, large batch sizes, and a distributional critic are important for achieving strong performance with off-policy RL algorithms.

\begin{figure*}[t!]
\vspace{-0.1in}
\centering
\subfloat[Effect of parallel environments]
{
\includegraphics[width=0.48\linewidth]{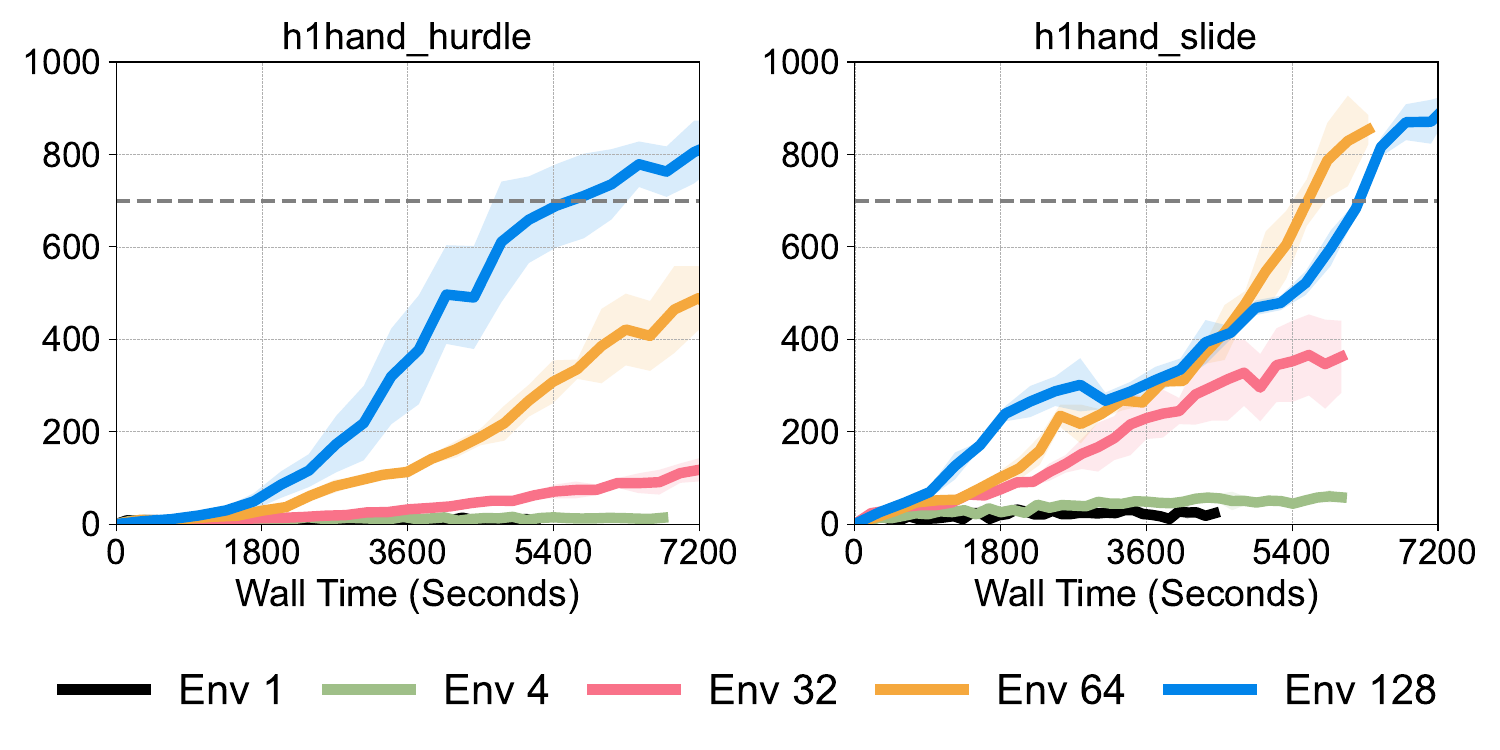}
\label{fig:analysis_num_envs}
}
\subfloat[Effect of batch size]
{
\includegraphics[width=0.48\linewidth]{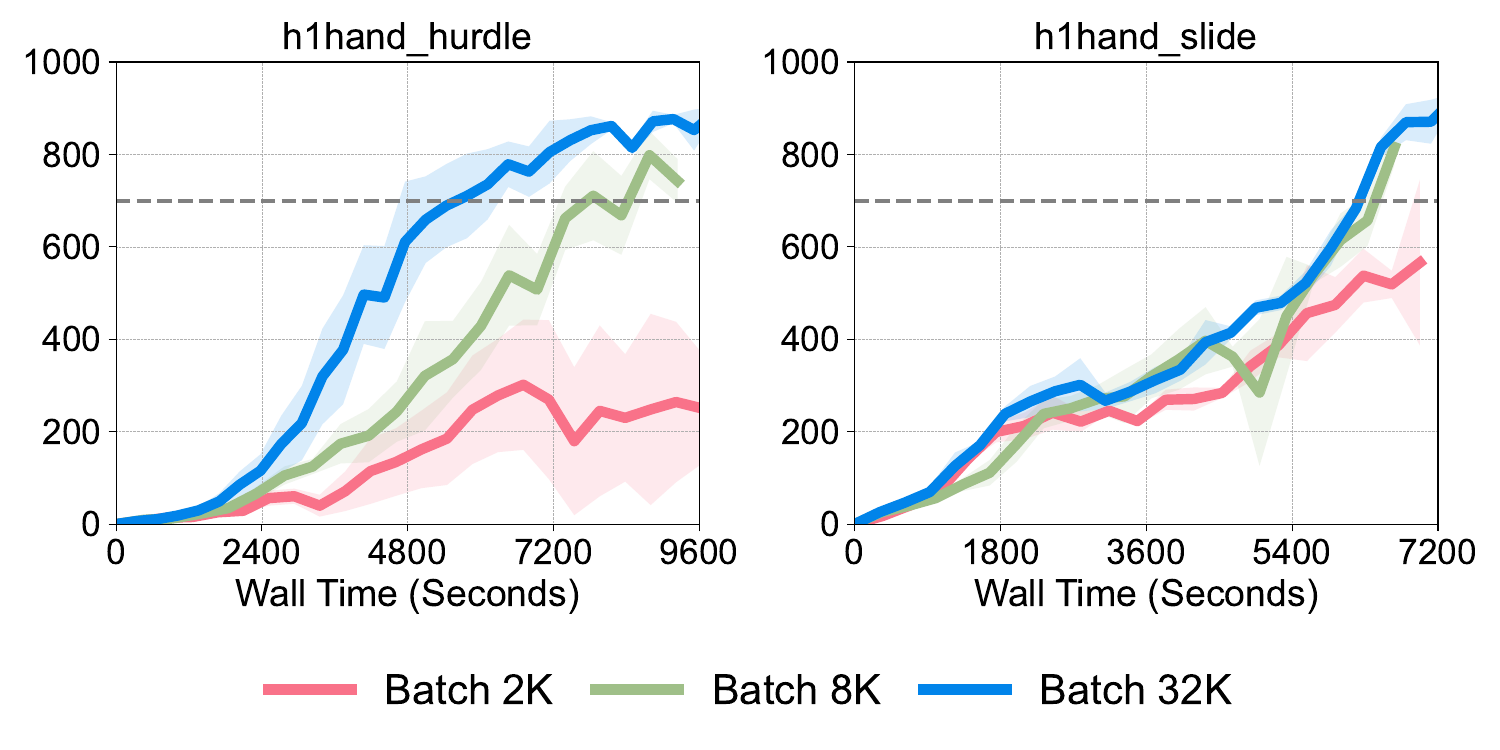}
\label{fig:analysis_batch_size}
}
\vspace{-0.125in}
\\
\subfloat[Effect of distributional RL]
{
\includegraphics[width=0.48\linewidth]{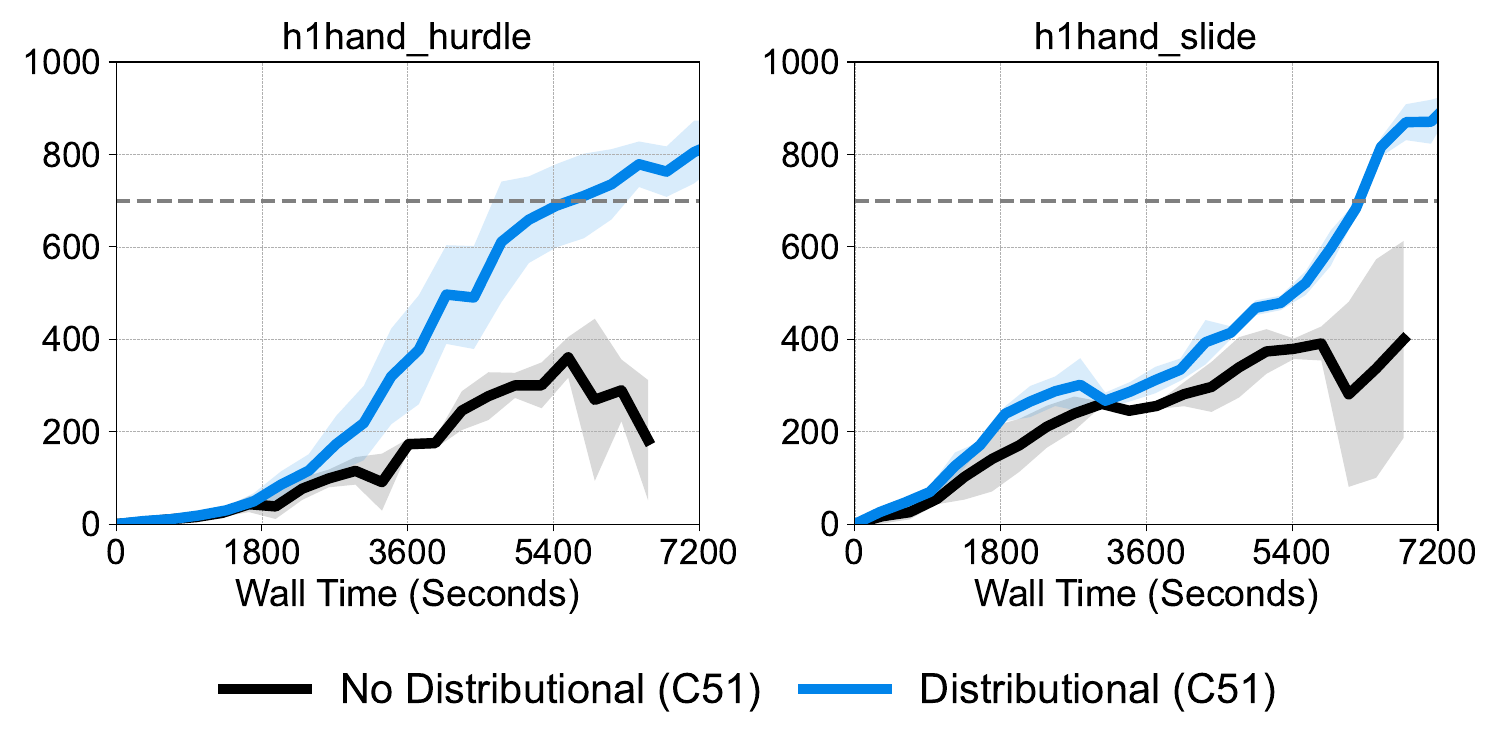}
\label{fig:analysis_c51}
}
\subfloat[Effect of Clipped Double Q-learning]
{
\includegraphics[width=0.48\linewidth]{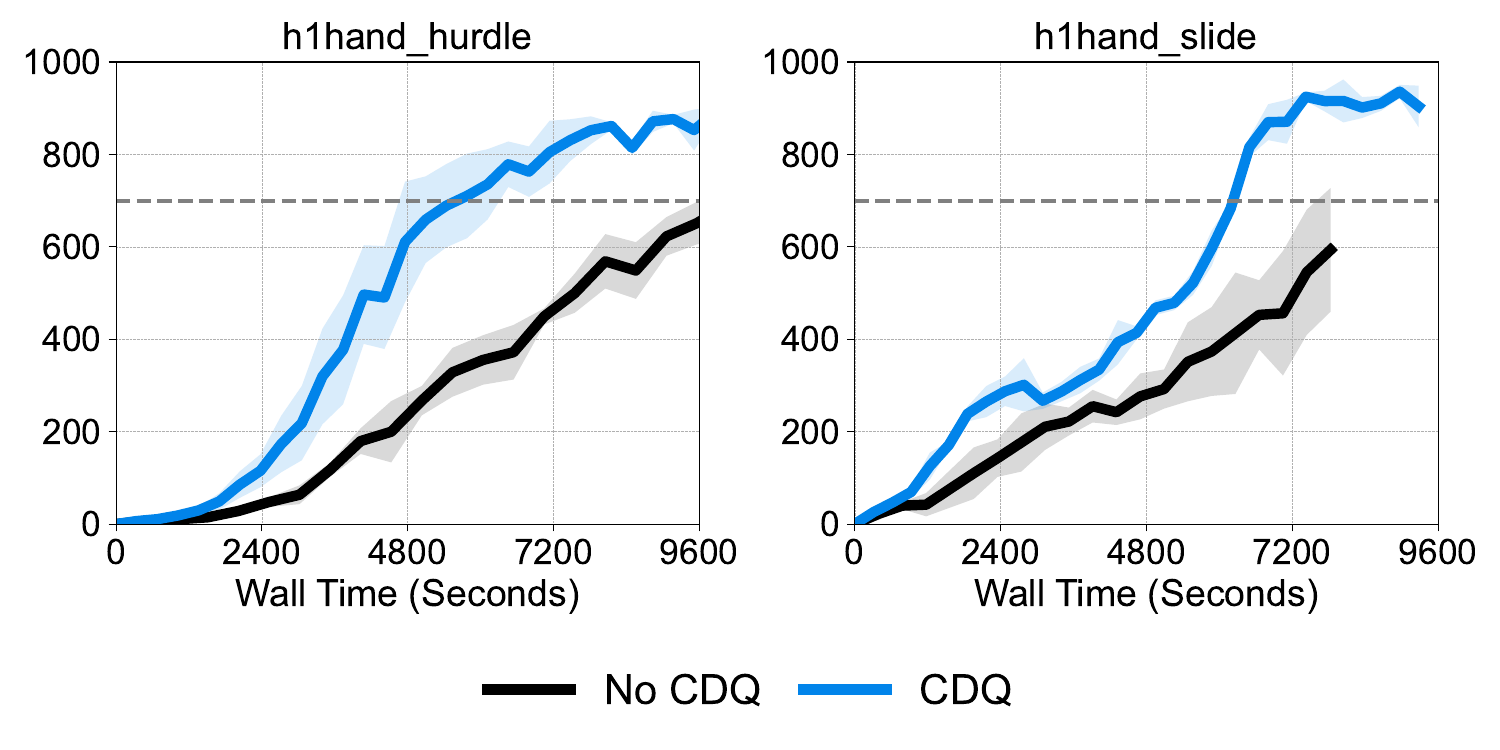}
\label{fig:analysis_cdq}
}
\caption{\textbf{Effect of design choices (1 / 2).} We investigate the effect of (a) parallel environments, (b) batch size, (c) distributional RL, and (d) Clipped Double Q-learning. The solid line and shaded regions represent the mean and standard deviation across three runs.
}
\label{fig:analysis1}
\vspace{-0.1in}
\end{figure*}

\vspace{-0.05in}
\subsection{Design Choices}
\label{sec:design_choices}
\vspace{-0.05in}

In this section, we describe the key design choices made in the development of FastTD3 and their impact on performance. For details of TD3, we refer the reader to \citet{fujimoto2018addressing}.

\vspace{-0.05in}
\paragraph{Parallel environments} Similar to observations in \citet{li2023mage}, we find that using massively parallel environments significantly accelerates TD3 training.
We hypothesize that combining deterministic policy gradient algorithms \citep{silver2014deterministic} with parallel simulation is particularly effective, because the randomness from parallel environments increases diversity in the data distribution.
This enables TD3 to leverage its strength -- efficient exploitation of value functions -- while mitigating its weakness in exploration.

\vspace{-0.05in}
\paragraph{Large-batch training}
We find that using an unusually large batch size of 32,768 for training the FastTD3 agent is highly effective.
We hypothesize that, with massively parallel environments, large-batch updates provide a more stable learning signal for the critic by ensuring high data diversity in each gradient update.
Otherwise, unless with high update-to-data ratio, a large portion of data will never be seen by the agent.
While increasing the batch size incurs a higher per-update wall-clock time, it often reduces overall training time due to improved training efficiency.

\vspace{-0.05in}
\paragraph{Distributional RL}
We also find that using the distributional critic \citep{bellemare2017distributional} is helpful in most cases, similar to the observation of \citet{li2023parallel}.
However, we note that this comes at the cost of additional hyperparameters -- $v_{\tt{min}}$ and $v_{\tt{max}}$.
Although we empirically find that they are not particularly difficult to tune\footnote{We provide tuned hyperparameters in our open-source implementation.}, one may be able to consider incorporating the reward normalization for the distributional critic proposed in SimbaV2 \citep{lee2025hyperspherical} into FastTD3.

\vspace{-0.05in}
\paragraph{Clipped Double Q-learning (CDQ)}
While \citet{Nauman2024overestimation} report that using the average of Q-values rather than the minimum employed in CDQ leads to better performance when combined with layer normalization, our findings indicate a different trend in the absence of layer normalization. Specifically, without layer normalization, CDQ remains a critical design choice and using minimum generally performs better across a range of tasks. This suggests that CDQ continues to be an important hyperparameter that must be tuned per task to achieve optimal reinforcement learning performance.

\vspace{-0.05in}
\paragraph{Architecture}
We use an MLP with a descending hidden layer configuration of 1024, 512, and 256 units for the critic, and 512, 256, and 128 units for the actor.
We find that using smaller models tends to degrade both time-efficiency and sample-efficiency in our experiments.
We also experiments with residual paths and layer normalization \citep{ba2016layer} similar to BRO \citep{nauman2024bigger} or Simba \citep{lee2024simba}, but they tend to slow down training without significant gains in our experiments.
We hypothesize that this is because the data diversity afforded by parallel simulation and large-batch training reduces the effective off-policyness of updates, thereby mitigating instability often associated with the deadly triad of bootstrapping, function approximation, and off-policy learning \citep{sutton2018reinforcement}. As a result, the training process remains stable even without additional architectural stabilizers like residual connections or layer normalization.

\vspace{-0.05in}
\paragraph{Exploration noise schedules}
In contrast to PQL \citep{li2023parallel} which found the effectiveness of mixed noise -- using different Gaussian noise scales for each environment sampled from $[\sigma_{\tt{min}}, \sigma_{\tt{max}}]$, we find no significant gains from the mixed noise scheme.
Nonetheless, we used mixed noise schedule, as it allows for flexible noise scheduling with only a few lines of additional code.
But we find that using large $\sigma_{\tt{max}} = 0.4$ is helpful for FastTD3 as shown in \citet{li2023parallel}.

\vspace{-0.05in}
\paragraph{Update-to-data ratio}
In contrast to prior work showing that increasing the update-to-data (UTD) ratio -- that is, the number of gradient updates per environment step -- typically requires additional techniques \citep{d'oro2023sampleefficient,schwarzer2023bigger} or architectural changes \citep{nauman2024bigger, lee2024simba, lee2025hyperspherical}, we find that FastTD3 does not require such modifications.
Using a standard 3-layer MLP without normalization, FastTD3 scales favorably with higher UTDs in terms of sample efficiency.
In particular, we find sample-efficiency tends to improve with higher UTDs, but at the cost of increased wall-time for training.
We hypothesize that this is because FastTD3 operates at extremely low UTDs -- typically ${2, 4, 8}$ updates per 128 to 4096 (parallel) environment steps -- reducing the risk of early overfitting often associated with high UTDs.

\vspace{-0.05in}
\paragraph{Replay buffer size}
Instead of defining a \textit{global} replay buffer size, we set the size as $N \times \texttt{num\_envs}$ (see \cref{sec:implementation_details} for more details on replay buffer design). In practice, we find that using a larger $N$ improves performance, though it comes at the cost of increased GPU memory usage, as we store entire buffer on the GPU.

\subsection{Implementation Details}
\label{sec:implementation_details}

\vspace{-0.05in}
\paragraph{Parallel environments} For IsaacLab and MuJoCo Playground, we use their native support for parallel simulation.
However, for HumanoidBench that does not support GPU-based parallelization, we use \texttt{SubprocVecEnv} from Stable Baselines3 library \citep{stable-baselines3}.
We find that HumanoidBench's default configuration launches a GPU-based renderer for each simulation, which makes it difficult to run more than 100 environments.
We have submitted a pull request to HumanoidBench that adds support for disabling the default GPU-based renderer, which is merged into the main branch.

\vspace{-0.05in}
\paragraph{Environment wrappers} To build an easy-to-use codebase that supports different suites that assume different configurations, we built or used wrappers for each suite.
\begin{itemize}
    \item For MuJoCo Playground, we use the native \texttt{RSLRLBraxWrapper} that converts Jax tensors to Torch tensors and follows the API of RSL-RL \citep{rudin2022learning}. Because this wrapper does not support saving \textit{final} observations before resetting each environment, we implemented this in a separate fork\footnote{See \url{https://github.com/younggyoseo/mujoco_playground}.}. We will get these changes to be merged into the main repository.
    \item As IsaacLab natively supports PyTorch, we implemented a simple wrapper that conforms to the RSL-RL API. Currently, our implementation does not support rendering during IsaacLab training, as IsaacLab does not allow for multiple simulations to run concurrently.
    \item For HumanoidBench, we developed a wrapper that follows the RSL-RL API and converts NumPy arrays to PyTorch tensors.
\end{itemize}

\vspace{-0.025in}
\paragraph{Asymmetric actor-critic} For IsaacLab and MuJoCo Playground, which often provide  privileged states for the critic network, we implement support for the asymmetric actor-critic \citep{pinto2017asymmetric}.

\vspace{-0.025in}
\paragraph{AMP and \texttt{torch.compile}}
While JAX-based RL implementations have become popular in recent days for its speed, we build our implementation upon PyTorch \citep{paszke2019pytorch} for its simplicity and flexibility.
We find that mixed-precision training with AMP and bfloat16 accelerates training by up to 40\% in our setup using a single A100 GPU, with no observed instability. We also support \texttt{torch.compile} by building our codebase on LeanRL \citep{huang2022cleanrl}\footnote{\url{https://github.com/pytorch-labs/LeanRL}}, which provides up to a 35\% speedup in our experiments. When using both AMP and \texttt{torch.compile}, we observe a combined training speedup of up to 70\%.

\paragraph{Replay buffer} Instead of defining a \textit{global} replay buffer size, we set the size as $N \times \texttt{num\_envs}$.
We find that it better decouples the effects of replay buffer size from the number of parallel environments.
For instance, if the global buffer size is fixed at one million, the task episode length is 1000, and the user increases the number of environments from 1000 to 2000, the replay buffer will only be able to save half of each trajectory from each environment (it will start discarding early samples after 500 timesteps), which may negatively affect performance.
On the other hand, if we specify $N=1000$ for each environment, the buffer can save the whole trajectory without being affected by the number of parallel environments.
Because we focus on non-vision domains, we store the entire buffer on the GPU to avoid the overhead of data transfer between CPU and GPU.

\begin{figure*}[t!]
\centering
\subfloat[Effect of model size]
{
\includegraphics[width=0.48\linewidth]{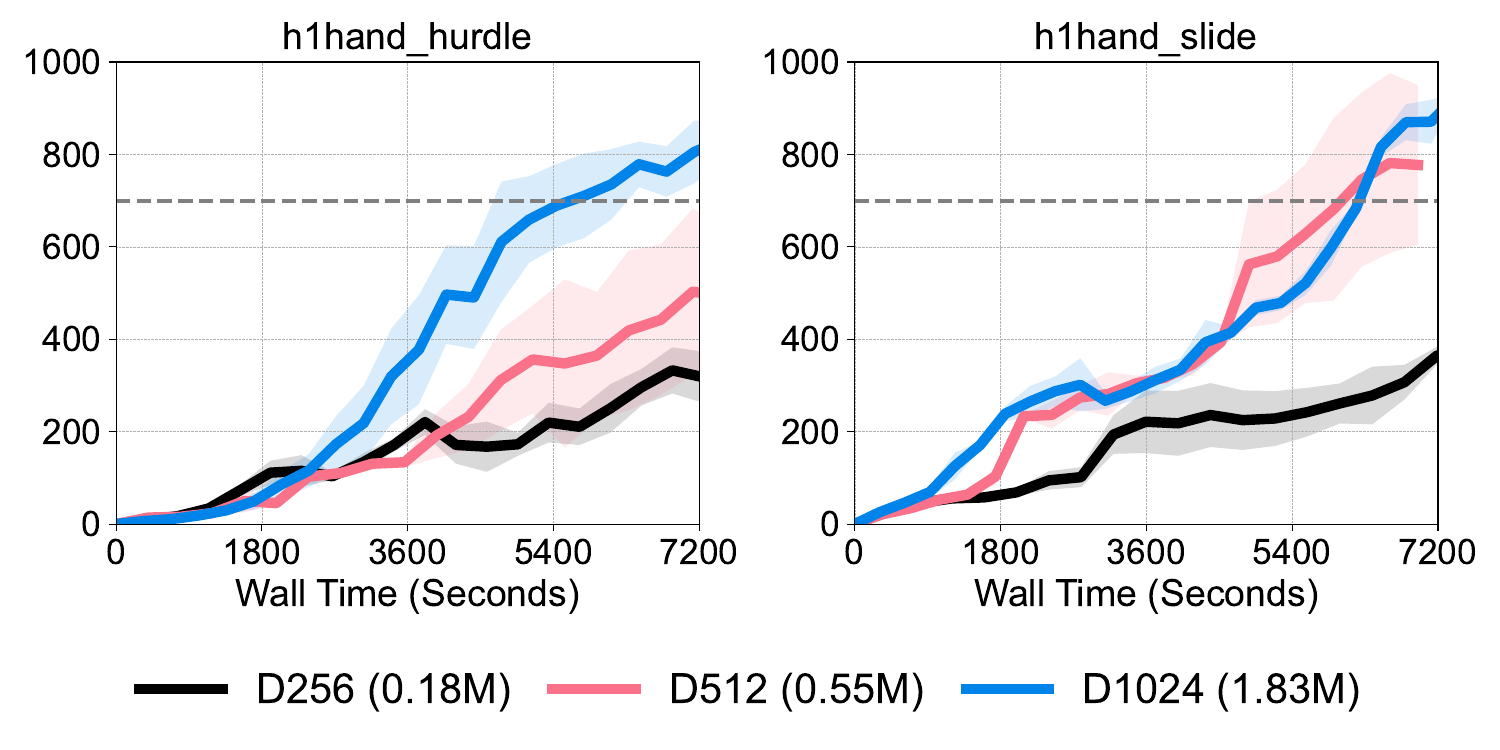}
\label{fig:analysis_model_size}
}
\subfloat[Effect of noise scales]
{
\includegraphics[width=0.48\linewidth]{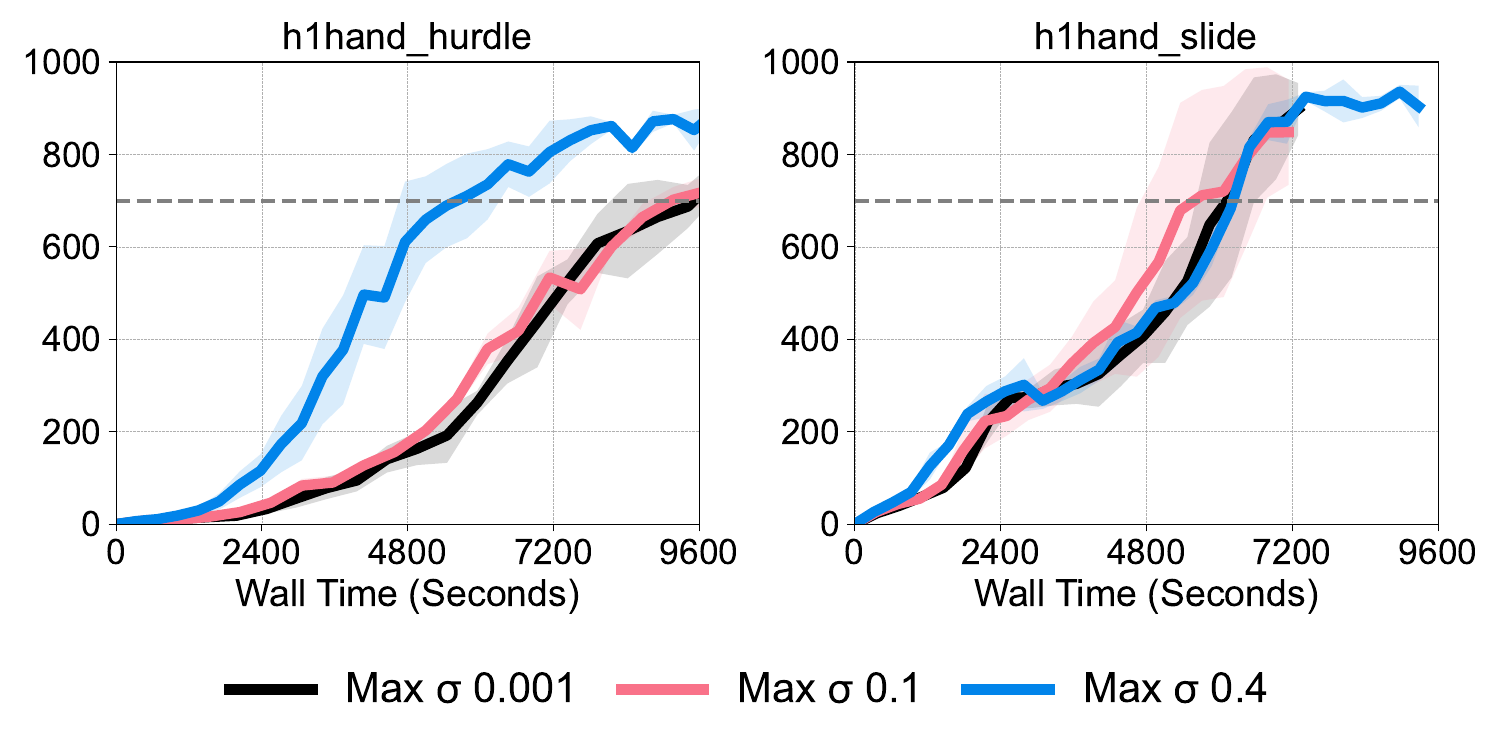}
\label{fig:analysis_max_sigma}
}
\vspace{-0.1in}
\\
\subfloat[Effect of update-to-data ratio]
{
\includegraphics[width=0.48\linewidth]{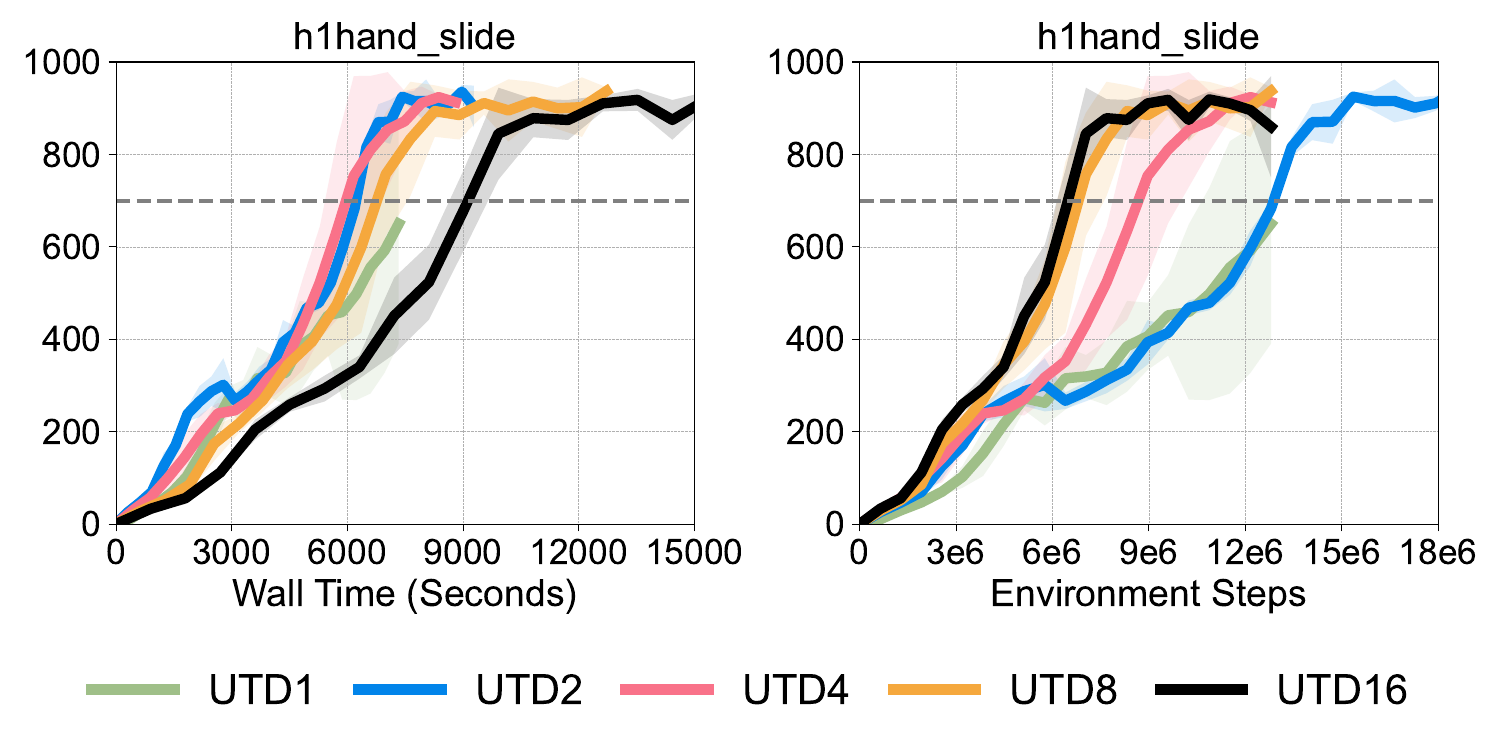}
\label{fig:analysis_utd}
}
\subfloat[Effect of replay buffer size]
{
\includegraphics[width=0.48\linewidth]{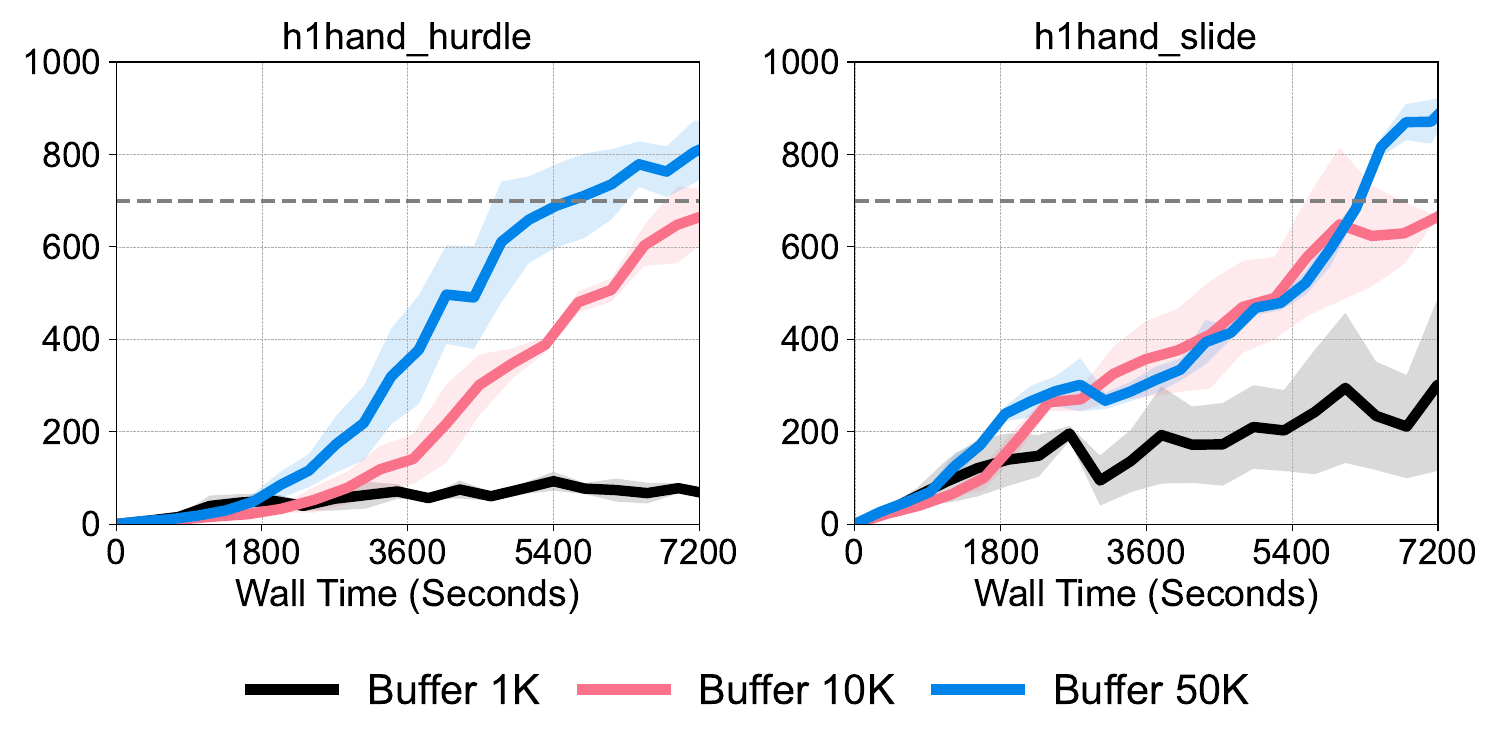}
\label{fig:analysis_buffer_size}
}
\caption{\textbf{Effect of design choices (2 / 2).} We investigate the effect of (a) model size, (b) noise scales, (c) update-to-data ratio, and (d) replay buffer size. The solid line and shaded regions represent the mean and standard deviation across three runs.
}
\vspace{-0.05in}
\label{fig:analysis2}
\end{figure*}

\begin{figure*}[t!]
\centering
\subfloat[
  Gait of 
  \sethlcolor{blue!15}\hl{FastTD3} trained with
  \sethlcolor{yellow!30}\hl{PPO}-tuned reward
]
{
\includegraphics[width=0.98\linewidth]{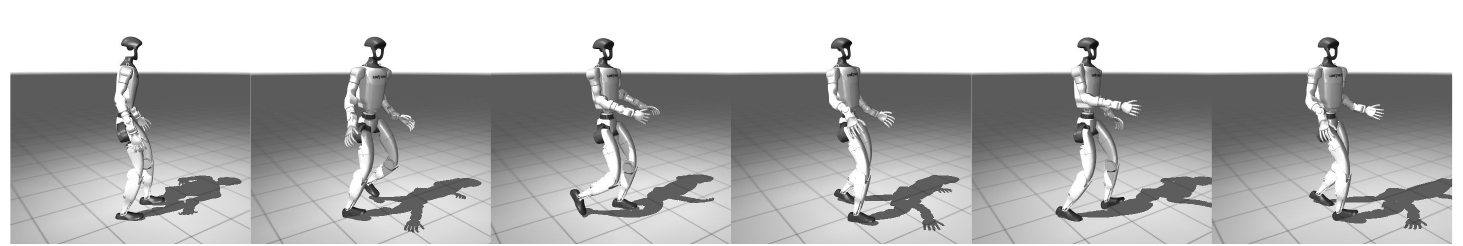}
\label{fig:analysis_gaits_fasttd3_with_ppo_reward}
}
\vspace{-0.15in}
\\
\subfloat[
  Gait of 
  \sethlcolor{yellow!30}\hl{PPO} trained with
  \sethlcolor{yellow!30}\hl{PPO}-tuned reward
]
{
\includegraphics[width=0.98\linewidth]{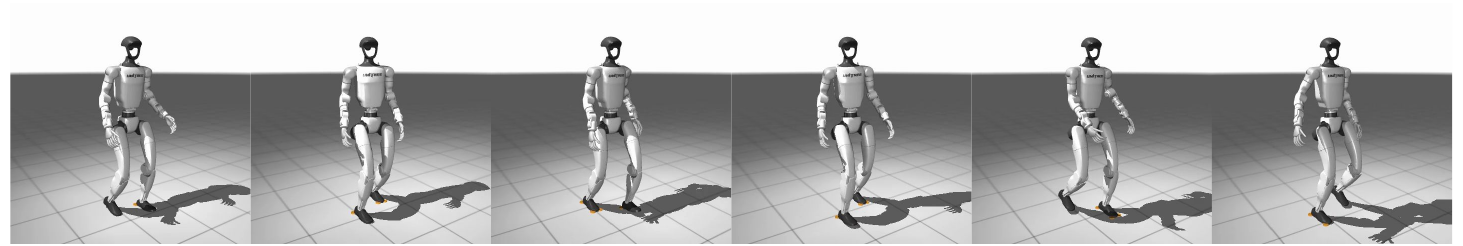}
\label{fig:analysis_gaits_ppo_with_ppo_reward}
}
\vspace{-0.15in}
\\
\subfloat[
  Gait of 
  \sethlcolor{blue!15}\hl{FastTD3} trained with
  \sethlcolor{blue!15}\hl{FastTD3}-tuned reward
]
{
\includegraphics[width=0.98\linewidth]{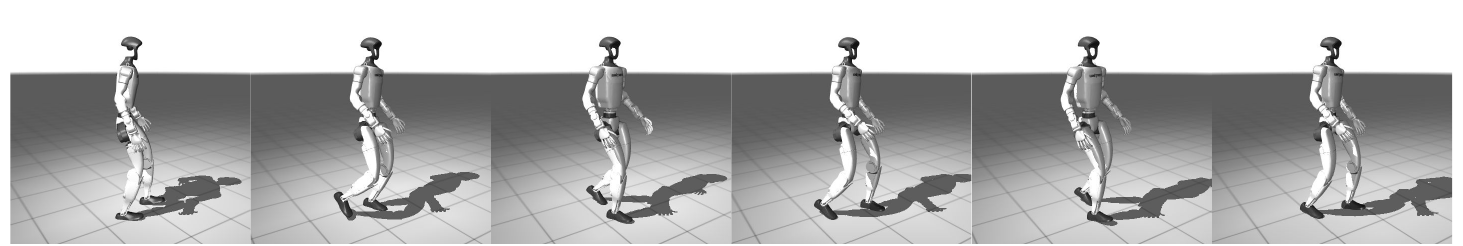}
\label{fig:analysis_gaits_fasttd3_with_fasttd3_reward}
}
\vspace{-0.15in}
\\
\subfloat[
  Gait of 
  \sethlcolor{yellow!30}\hl{PPO} trained with
  \sethlcolor{blue!15}\hl{FastTD3}-tuned reward
]
{
\includegraphics[width=0.98\linewidth]{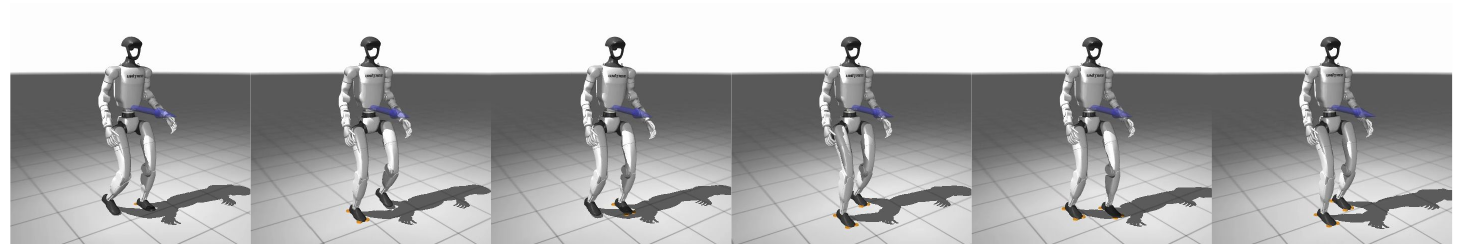}
\label{fig:analysis_gaits_ppo_with_fasttd3_reward}
}
\caption{\textbf{Different RL algorithms may need different reward functions.} We find that \sethlcolor{blue!15}\hl{FastTD3} trained with the \sethlcolor{yellow!30}\hl{PPO}-tuned reward function (a) results in an undeployable gait, as the G1 robot exhibits abrupt arm movements. (b) This is notable because the same (\sethlcolor{yellow!30}\hl{PPO}-tuned) reward function successfully produces a natural walking gait when used with \sethlcolor{yellow!30}\hl{PPO}.
(c) By tuning the reward function specifically for \sethlcolor{blue!15}\hl{FastTD3} -- adding stronger penalty terms -- we were able to train a smoother gait with \sethlcolor{blue!15}\hl{FastTD3}.
(d) On the other hand, training \sethlcolor{yellow!30}\hl{PPO} with \sethlcolor{blue!15}\hl{FastTD3}-tuned reward is slow to train and results in a slowly-walking gait because of stronger penalty terms.
For clarity, we use \sethlcolor{yellow!30}\hl{PPO}-tuned reward for all experiments in other experiments except this analysis and sim-to-real experimental results.
}
\label{fig:analysis_gaits}
\vspace{-0.05in}
\end{figure*}

\section{Experiments}

\vspace{-0.025in}
\paragraph{Setups}
For the DreamerV3 \citep{hafner2023mastering}, SAC \citep{haarnoja2018soft}, and TDMPC2 \citep{hansen2023td} baselines, we use learning curves from three runs available in the HumanoidBench repository.
Given that each run is trained for 48 hours, we use interpolated wall-clock timestamps to plot the curves.
For SimbaV2 \citep{lee2025hyperspherical}, we use the official codebase to conduct experiments on tasks involving dexterous hands\footnote{SimbaV2 uses action repeat of 2 for HumanoidBench in their paper, which is unusual for joint position control. We thus run our SimbaV2 experiments with action repeat of 1.}. 
We report single-seed results for SimbaV2 but plan to include additional runs in the future.
All FastTD3 results are aggregated over three runs. Experiments are conducted on a cloud instance with a single NVIDIA A100 GPU and 16 CPU cores.

\vspace{-0.025in}
\paragraph{Results}
We provide the aggregate results over all tasks for each suite in \cref{fig:aggregate}, individual results on a selected set of tasks in \cref{fig:aggregate_total}, and individual results on full set of tasks in \cref{appendix:additional_results}.
We provide extensive experimental results that investigate the effect of various design choices (described in \cref{sec:design_choices}) in \cref{fig:analysis1} and \cref{fig:analysis2}.

\vspace{-0.025in}
\paragraph{Different RL algorithms may need different reward functions}
While training humanoid locomotion policies in MuJoCo Playground, we observed that PPO and FastTD3 produced notably different gaits, despite being trained with the same reward function (see \cref{fig:analysis_gaits_fasttd3_with_ppo_reward} and \cref{fig:analysis_gaits_ppo_with_ppo_reward}).
We hypothesize that this is because existing reward functions are typically tuned for PPO, and different algorithms may require different reward structures to produce desirable behaviors.
To address this, we tuned the reward function specifically for FastTD3 -- with stronger penalty terms. This process was efficient due to FastTD3’s short training time. As shown in \cref{fig:analysis_gaits_fasttd3_with_fasttd3_reward}, the tuned reward enabled FastTD3 to learn a stable and visually appealing gait compared to the one in \cref{fig:analysis_gaits_fasttd3_with_ppo_reward}.
On the other hand, we observe that training PPO with FastTD3-tuned reward results also results in undeployable gait that walks too slowly (see \cref{fig:analysis_gaits_ppo_with_fasttd3_reward}).
This observation suggests that standard metric -- episode return -- may not capture the practical usefulness of learned policies.

\begin{figure*} [t!] \centering
\includegraphics[width=0.99\textwidth]{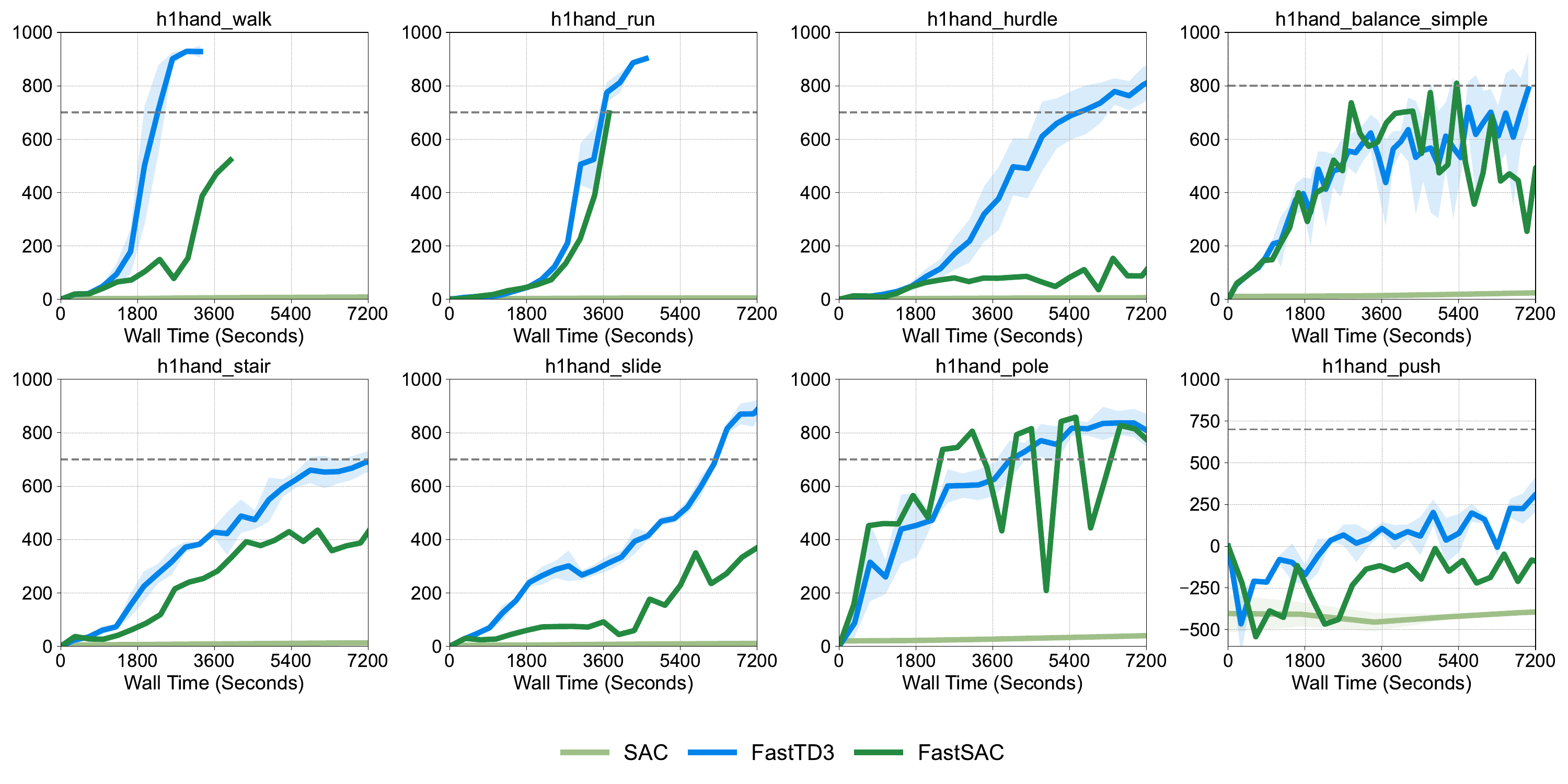}
\vspace{-0.05in}
\caption{\textbf{FastSAC.} We develop FastSAC, a variant of SAC that incorporates our FastTD3 recipe. We find that FastSAC is significantly faster than vanilla SAC, though still slower than FastTD3.}
\label{fig:comparison_fastsac}
\vspace{-0.05in}
\end{figure*}

\paragraph{FastSAC Experiments}
To investigate whether our recipe generalizes to other model-free RL algorithms, we develop FastSAC, which incorporates our FastTD3 recipe into SAC \citep{haarnoja2018soft}. We find that FastSAC trains significantly faster than vanilla SAC (see \cref{fig:comparison_fastsac}). However, we also observe that FastSAC tends to be unstable during training, which we hypothesize is due to the difficulty of maximizing action entropy in high-dimensional action spaces.
Given that SimbaV2 \citep{lee2025hyperspherical} is notably faster than vanilla SAC in our main experiments, incorporating such recent advancements in off-policy RL into FastTD3 or FastSAC may be a promising future direction.

\paragraph{Sim-to-real RL with FastTD3} 
For the sim-to-real experiments in \cref{fig:sim2real}, we use Booster Gym\footnote{\url{https://github.com/BoosterRobotics/booster_gym}}, which supports 12-DOF control for a Booster T1 humanoid robot with fixed arms, waist, and heading.
For convenience, we ported Booster Gym's robot configuration and reward functions into MuJoCo Playground, which originally only supports 23-DOF T1 control with all joints enabled.
We find that training FastTD3 in MuJoCo Playground significantly simplifies and accelerates iteration cycle compared to training with Booster Gym based on IsaacGym \citep{makoviychuk2021isaac}.

\section{Discussion}
We have presented FastTD3, a simple, fast, and capable RL algorithm that efficiently solves a variety of locomotion and manipulation tasks from HumanoidBench \citep{sferrazza2024humanoidbench}, IsaacLab \citep{mittal2023orbit}, and MuJoCo Playground \citep{zakka2025mujoco}.
We have demonstrated that a simple algorithm, combined with well-tuned hyperparameters and without introducing new architectures or training techniques, can serve as a surprisingly strong baseline for complex robotics tasks.
Along with this report, we provide an open-source implementation of FastTD3 -- a lightweight and easy-to-use codebase featuring user-friendly features and pre-configured hyperparameters.

Here, we would like to emphasize the goal of this work is not to claim novelty or superiority over prior algorithms. 
Our approach builds directly on insights already established in the research community. Parallel Q-Learning (PQL; \citealt{li2023parallel}) demonstrated how off-policy RL can be effectively scaled using massive parallel simulation, followed by Parallel Q-Network (PQN; \citealt{gallici2024simplifying}) that made a similar observation for discrete control.
Similarly, \citet{raffin2025isaacsim} and \citet{shukla2025fastsac} showed that SAC can also be scaled successfully through parallel simulation and carefully tuned hyperparameters.
The aim of this work is to distill those insights into a simple algorithm, provide extensive experimental analysis of various design choices, and release an easy-to-use implementation.

We are excited about several future directions.
Importantly, our work is orthogonal to many recent advances in RL, and these improvements can be readily incorporated into FastTD3 to further advance the state of the art.
We expect this integration process to be both straightforward and effective.
We also look forward to applications of FastTD3 in real-world RL setups. 
As an off-policy RL algorithm, FastTD3 is well-suited for demo-driven RL setups for humanoid control \citep{chernyadev2024bigym,seo2024reinforcement}, as well as for fine-tuning simulation-trained policies through real-world interactions.
Finally, the faster iteration cycles of FastTD3 could be useful in iterative inverse RL setups that leverage language models as reward generators \citep{ma2023eureka}, offering a promising approach to address the longstanding challenge of reward design in humanoid control.

We hope that our work and implementation help accelerate future RL research in robotics.

\section*{Acknowledgements}
This work is supported in part by Multidisciplinary University Research Initiative (MURI) award by the Army Research Office (ARO) grant No. W911NF-23-1-0277 and the ONR Science of Autonomy Program N000142212121, and ONR MURI N00014-22-1-2773.
Pieter Abbeel holds concurrent appointments as a Professor at UC Berkeley and as an Amazon Scholar. This paper describes work performed at UC Berkeley and is not associated with Amazon.
We thank NVIDIA for providing
compute resources through the NVIDIA Academic DGX Grant.

\bibliography{reference}
\bibliographystyle{ref_bst}

\newpage
\appendix
\onecolumn

\section{Additional Results}
\label{appendix:additional_results}

\subsection{HumanoidBench}
\label{appendix:humanoid_bench}

\begin{figure*} [h] \centering
\includegraphics[width=0.99\textwidth]{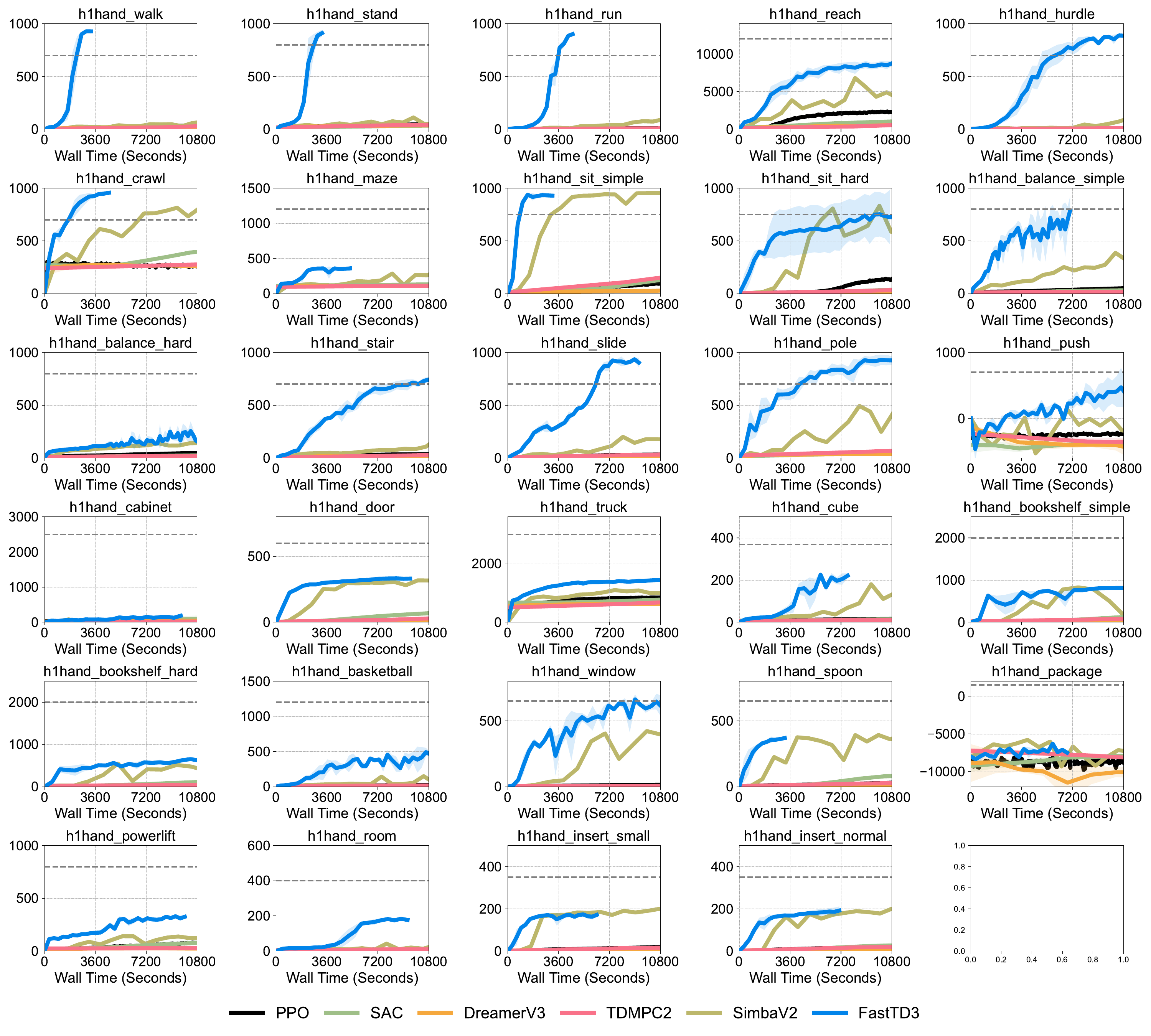}
\caption{\textbf{HumanoidBench results.} We provide learning curves on a 39 tasks from HumanoidBench \citep{sferrazza2024humanoidbench} The solid line and shaded regions represent the mean and standard deviation across three runs.}
\label{fig:full_humanoid_bench}
\vspace{-0.05in}
\end{figure*}

\clearpage

\subsection{IsaacLab}
\label{appendix:isaaclab}

\begin{figure*} [h] \centering
\includegraphics[width=0.99\textwidth]{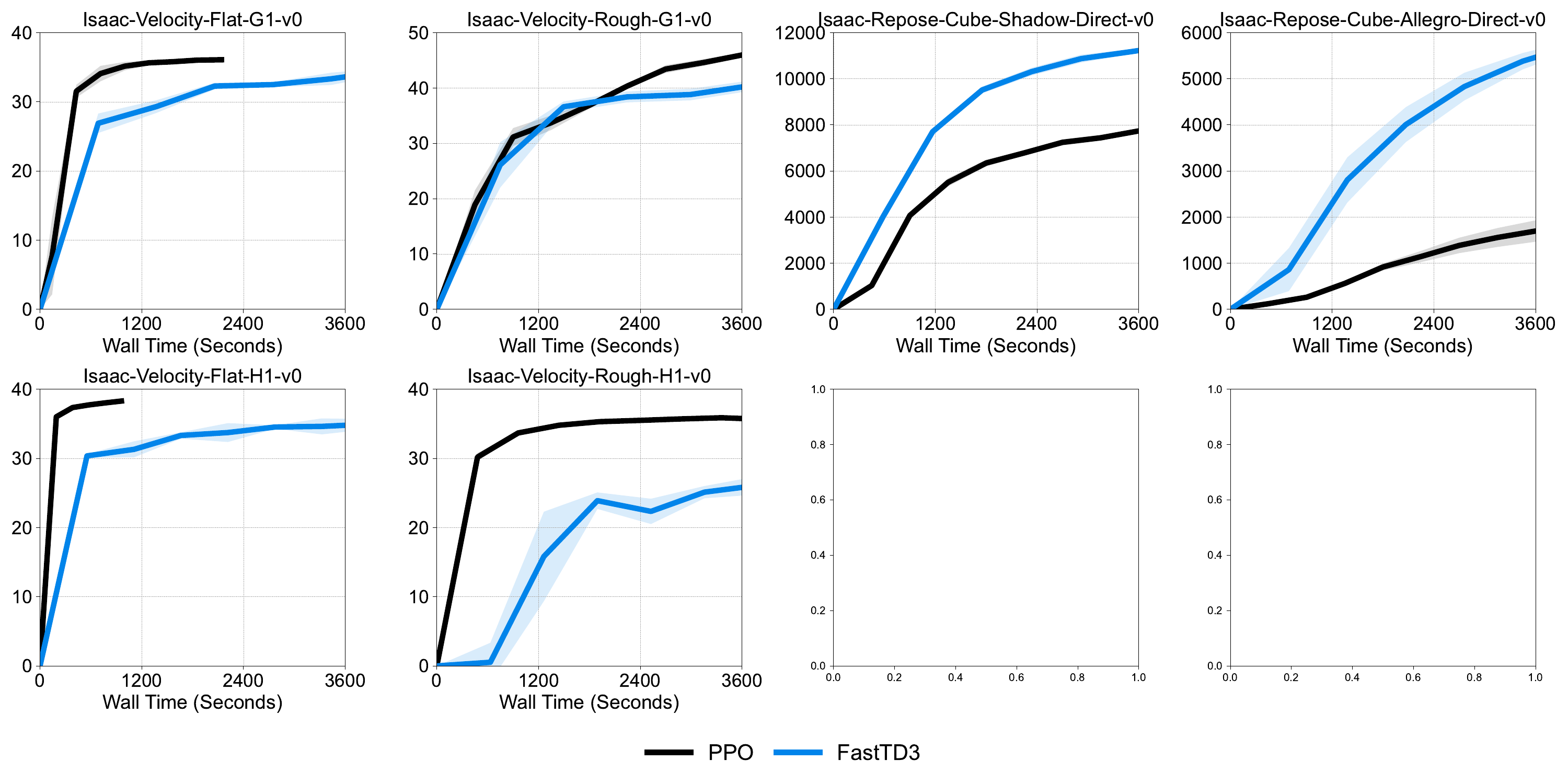}
\caption{\textbf{IsaacLab results.} We provide learning curves on six tasks from IsaacLab \citep{mittal2023orbit}. The solid line and shaded regions represent the mean and standard deviation across three runs.}
\label{fig:full_isaaclab}
\vspace{-0.05in}
\end{figure*}

\subsection{MuJoCo Playground}
\label{appendix:playground}

\begin{figure*} [h] \centering
\includegraphics[width=0.99\textwidth]{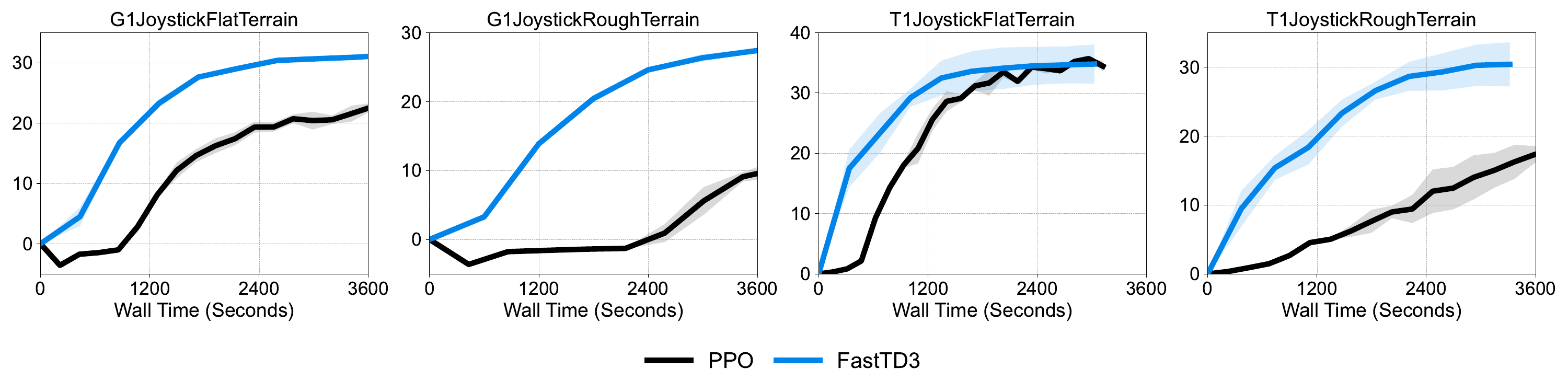}
\caption{\textbf{MuJoCo Playground results.} We provide learning curves on four tasks from MuJoCo Playground \citep{zakka2025mujoco}. The solid line and shaded regions represent the mean and standard deviation across three runs.}
\label{fig:full_playground}
\vspace{-0.05in}
\end{figure*}

\clearpage

\section{Sample Efficiency Curves}

\subsection{HumanoidBench}
\label{appendix:humanoid_bench_sample_efficiency}

\begin{figure*} [h] \centering
\includegraphics[width=0.99\textwidth]{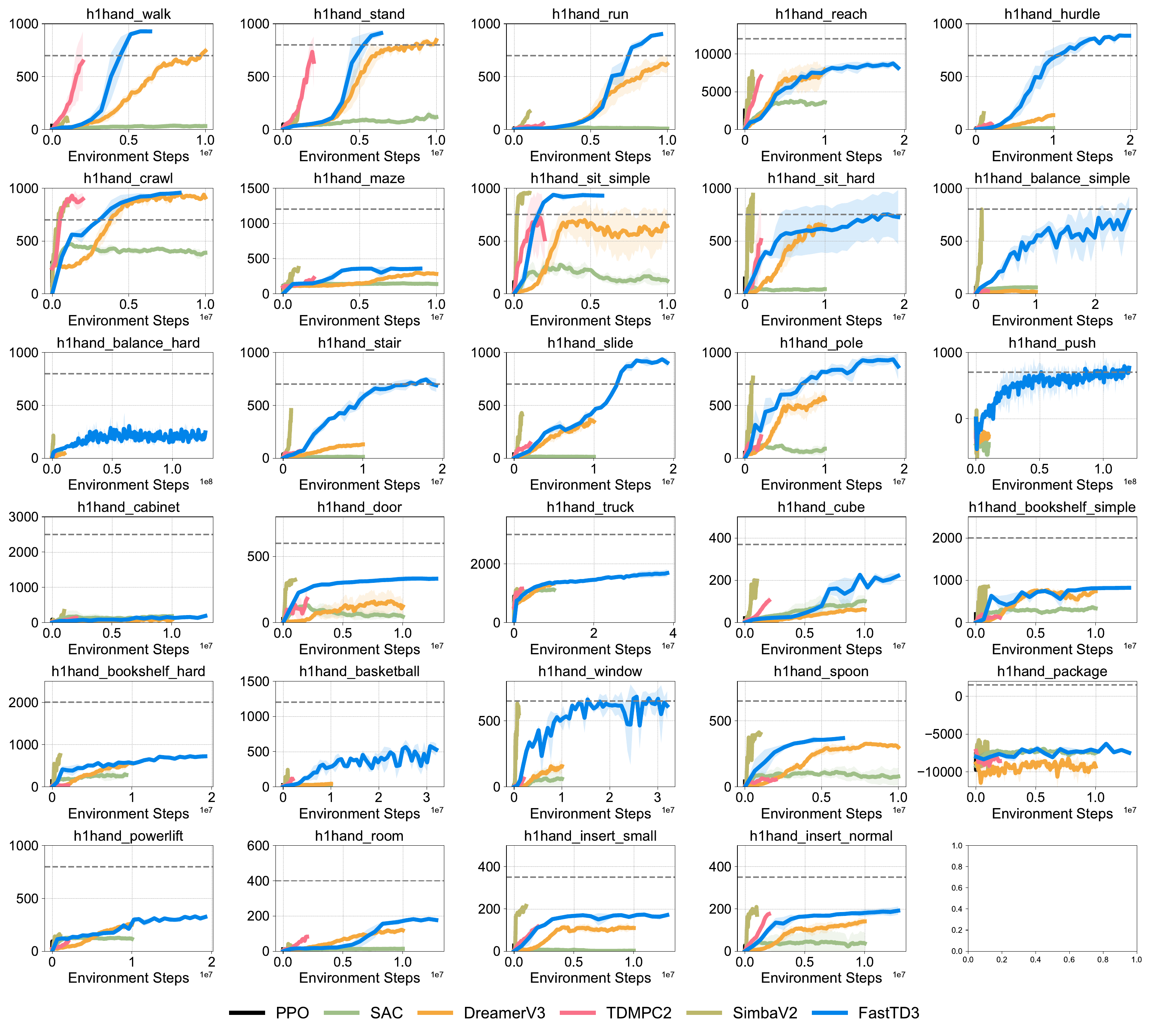}
\caption{\textbf{HumanoidBench results (sample-efficiency).} We provide learning curves on a 39 tasks from HumanoidBench \citep{sferrazza2024humanoidbench} The solid line and shaded regions represent the mean and standard deviation across three runs.}
\label{fig:full_humanoid_bench_sample_efficiency}
\vspace{-0.05in}
\end{figure*}

\clearpage

\subsection{IsaacLab}
\label{appendix:isaaclab_sample_efficiency}

\begin{figure*} [h] \centering
\includegraphics[width=0.99\textwidth]{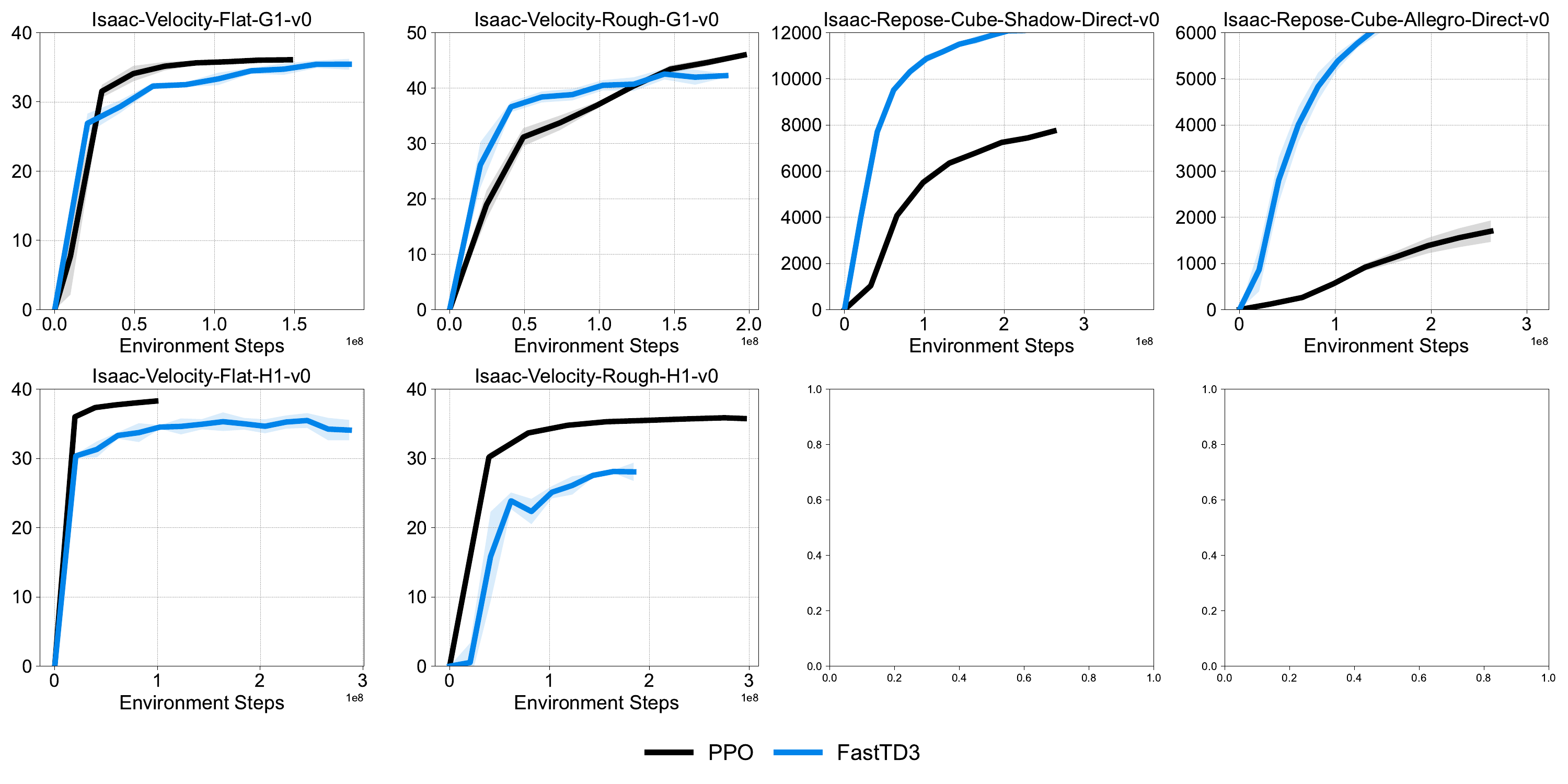}
\caption{\textbf{IsaacLab results (sample-efficiency).} We provide learning curves on six tasks from IsaacLab \citep{mittal2023orbit}. The solid line and shaded regions represent the mean and standard deviation across three runs.}
\label{fig:full_isaaclab_sample_efficiency}
\vspace{-0.05in}
\end{figure*}

\subsection{MuJoCo Playground}
\label{appendix:playground_sample_efficiency}

\begin{figure*} [h] \centering
\includegraphics[width=0.99\textwidth]{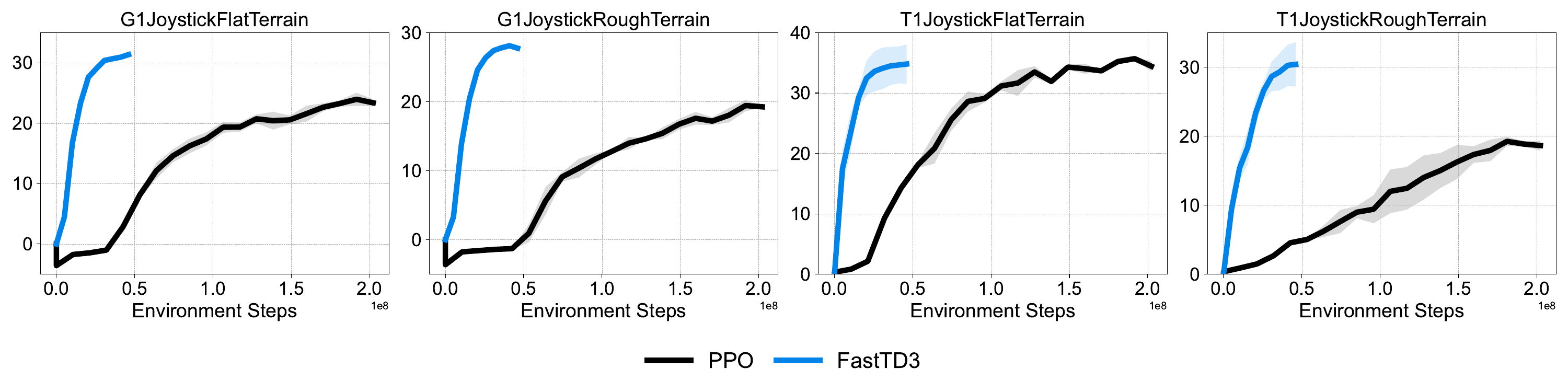}
\caption{\textbf{MuJoCo Playground results (sample-efficiency).} We provide learning curves on four tasks from MuJoCo Playground \citep{zakka2025mujoco}. The solid line and shaded regions represent the mean and standard deviation across three runs.}
\label{fig:full_playground_sample_efficiency}
\vspace{-0.05in}
\end{figure*}

\end{document}